\documentclass[runningheads]{llncs}
\usepackage{graphicx}

\usepackage{tikz}
\usepackage{comment}
\usepackage{amsmath,amssymb} 
\usepackage{color}
\usepackage{hyperref}
\usepackage{booktabs}
\usepackage{multirow}
\usepackage{marvosym}
\usepackage[accsupp]{axessibility}  
\definecolor{RURI}{RGB}{0 ,92,195}
\newcommand{\weather}[1]{\textcolor{RURI}{#1}}
\definecolor{AKABENI}{RGB}{203 , 64, 66}
\newcommand{\region}[1]{\textcolor{AKABENI}{#1}}
\definecolor{KIKYO}{RGB}{106 , 176, 56}
\newcommand{\scenetype}[1]{\textcolor{KIKYO}{#1}}

\begin{document}
\pagestyle{headings}
\mainmatter
\def\ECCVSubNumber{1277}  

\title{Learning to Drive by Watching YouTube Videos: Action-Conditioned Contrastive\\ Policy Pretraining} 

\titlerunning{Action-Conditioned Contrastive Policy Pretraining}
%
\author{Qihang Zhang\inst{1} \and
Zhenghao Peng\inst{1} \and
Bolei Zhou\inst{2}}
\authorrunning{Q. Zhang, Z. Peng, B. Zhou}
%
\institute{The Chinese University of Hong Kong, Hong Kong SAR, China \and
University of California, Los Angeles, USA \\
}
\maketitle

\begin{abstract}
Deep visuomotor policy learning, which aims to map raw visual observation to action, achieves promising results in control tasks such as robotic manipulation and autonomous driving. However, it requires a huge number of online interactions with the training environment, which limits its real-world application. Compared to the popular unsupervised feature learning for visual recognition, feature pretraining for visuomotor control tasks is much less explored. In this work, we aim to pretrain policy representations for driving tasks by watching hours-long uncurated YouTube videos. Specifically, we train an inverse dynamic model with a small amount of labeled data and use it to predict action labels for all the YouTube video frames. A new contrastive policy pretraining method is then developed to learn action-conditioned features from the video frames with pseudo action labels. Experiments show that the resulting action-conditioned features obtain substantial improvements for the downstream reinforcement learning and imitation learning tasks, outperforming the weights pretrained from previous unsupervised learning methods and ImageNet pretrained weight. 
Code, model weights, and data are available at: \href{https://metadriverse.github.io/ACO/}{https://metadriverse.github.io/ACO}.

\keywords{Learning From Unlabeled Data, Feature Pretraining}
\end{abstract}
\section{Introduction}

Deep policy learning makes promising progress to many visuomotor control tasks ranging from robotic manipulation~\cite{levine2016end,kalashnikov2018scalable,lange2012autonomous,wu2021example} to autonomous driving~\cite{chen2020learning,zhang2021roach}. 
By learning to map visual observation directly to control action through a deep neural network, it mitigates the manual design of controller, lowers the system complexity, and improves generalizability. However, the sample efficiency of the underlying algorithms such as reinforcement learning or imitation learning remains low. It requires a significant amount of online interactions or expert demonstrations 
in the training environment thus limits its real-world applications. 

\begin{figure}[!t]
\centering
\includegraphics[width=1\linewidth]{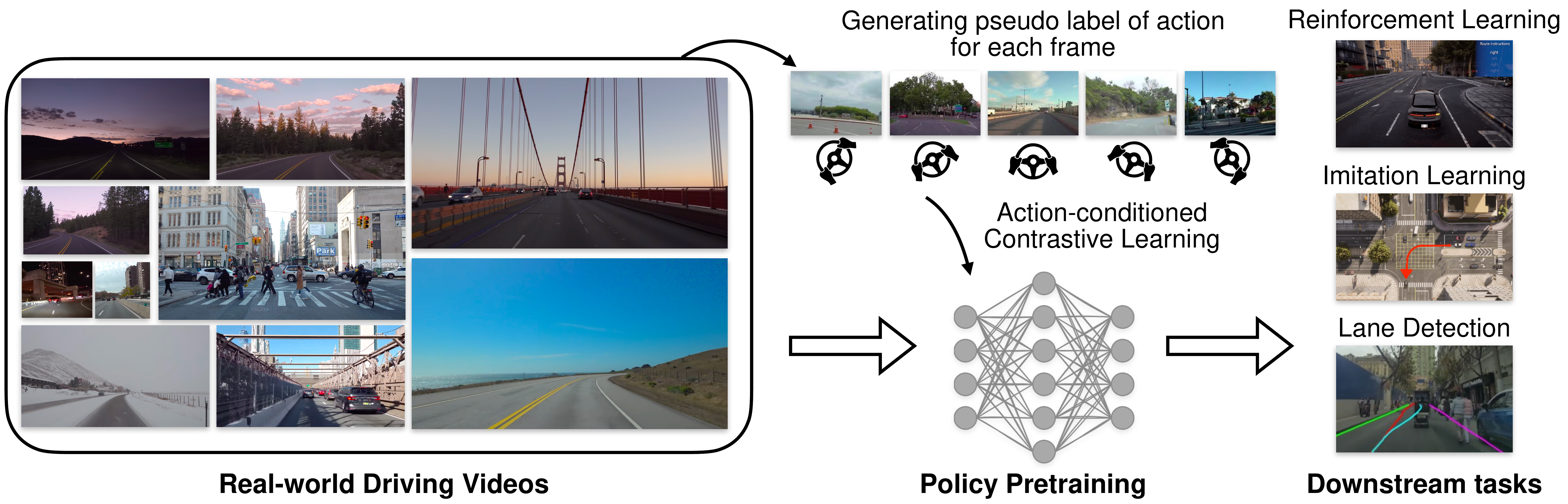}
\caption{
Hours-long YouTube videos contain a huge variety of driving scenes.
By developing an action-conditioned contrastive learning algorithm, our work aims to pretrain the visual representations with pseudo action labels on the diverse real-world data and improve sample efficiency, feature generalizability, and performance of the downstream tasks
}
\label{figure:teasor}
\end{figure}

Many recent works use unsupervised learning and data augmentation to improve the sample efficiency by pretraining the neural representations before policy learning. For example, random background videos are incorporated in the policy feature pretraining~\cite{hansen2020self,yarats2020image,hansen2021generalization}. 
However, the augmented data with random background videos
shifts drastically from the original data distribution,
which degrades the overall performance of the model.
Also, it remains challenging to generalize the learned weights to real world as it is hard to design augmentations that reflect the real-world diversity. 
In this work, we explore pretraining the neural representation on massive amount of real-world data directly. Figure~\ref{figure:teasor} shows some uncurated YouTube videos, which contain driving scenes all over the world with diverse conditions such as different weathers, urban and rural environments, and various traffic densities. We show that exploiting such real-world data in deep policy learning can substantially improve the generalizability of the learned weight and benefit various downstream tasks.

Learning deep representations from unlabeled data is a popular topic in visual recognition.
The learned representations are shown to be generalizable across visual tasks ranging from image classification, semantic segmentation, to object detection~\cite{he2020momentum,yang2021instance,grill2020bootstrap,chen2020simple,wu2018unsupervised}. 
However, these methods are mainly designed for learning visual features for recognition tasks, which are different from control tasks where an agent takes actions in uncertain environment. In control tasks, visual information such as the abstraction of appearance and texture of the scene may not be useful for decision making. For example, visual elements like lighting and weather are usually irrelevant to the driving task and might even become confounders in policy learning, negatively affecting the driving performance~\cite{zhang2020learning}. 
On the contrary, it is crucial to learn the features that matter to the output action.
For example, at the driving junction, the traffic light and the driving lane occupy only a few pixels in the visual observation but has a significant impact on the driver's actions.

In this paper, we propose a novel action-conditioned policy pretraining paradigm, which learns to capture important features in the neural representation relevant to decision making and benefits downstream tasks.
As shown in Figure~\ref{figure:teasor}, we first collect a large corpus of driving videos from YouTube which are recorded in 68 cities all over the world with a wide range of visual appearances.
We then train an inverse dynamics model with a small amount of labeled data and use it to  
generate pseudo action labels for each video frame. 
We finally develop a novel policy pretraining method called \textbf{A}ction-conditioned \textbf{CO}ntrastive Learning (\textbf{ACO}) that incorporates the action information in the representation learning. 
The motivation is to learn the discriminative features of video frames most related to driving actions. 
Specifically, instead of contrasting images based on different augmented views~\cite{he2020momentum}, we define a new contrastive pair conditioned on action similarity. 
By learning with those action-conditioned contrastive pairs, the representation captures visual elements that are highly correlated to the output actions. 

We evaluate the effectiveness of the action-conditioned pretraining for a variety of tasks, such as policy learning through Imitation Learning (IL) and Reinforcement Learning (RL) in end-to-end autonomous driving, and Lane Detection (LD).
The experimental results show that ACO successfully learns generalizable features for the downstream tasks.
Our contributions are summarized below:
\begin{enumerate}
    \item We propose a new paradigm of policy pretraining on a massive amount of real-world driving videos.
    \item We develop a novel action-conditioned contrastive learning approach ACO to learn action-related features.
    \item Experiments of various pretraining methods in downstream policy learning tasks show that the feature resulting from the proposed method achieves sufficient performance gain in the driving tasks. 
\end{enumerate}
\section{Related work}
\subsection{Image-based Policy Learning}
\noindent{\textbf{Learning with auxiliary task.}}
Reconstruction tasks~\cite{yarats2019improving,hafner2019learning,lee2020stochastic}, self-supervised objectives~\cite{laskin2020curl,stooke2021decoupling}, and future prediction objectives~\cite{hafner2019learning,yan2020learning,finn2015learning,pinto2016curious,agrawal2016learning} are proposed to mitigate the gap between state-based and image-based Reinforcement Learning (RL).
However, unsupervised learning based on visual information may not lead to optimal policy learning performance since the irrelevant visual features might become confounders in the representation and distract policy learning~\cite{zhang2020learning}.

\noindent{\textbf{Augmenting the input distribution.}}
Data augmentations like random shift~\cite{yarats2020image,hansen2021generalization} and inserting random videos at background~\cite{hansen2021generalization,hansen2020self} can improve policy robustness and generalization ability.
Visual and physical attributes of the environment and the agent can be randomized through domain randomization~\cite{tobin2017domain,andrychowicz2020learning,peng2018sim}.
However, it is hard to design augmentations that cover real world's variations. Significant computational resources are also required to train the policy with augmented input~\cite{tobin2017domain}.

\noindent{\textbf{Decoupling visual-level and policy-level learning.}}
Chen~\textit{et al.}~\cite{chen2020learning} first train a policy with full state observation and then use this cheating policy as a teacher to train an image-based policy. Similar teacher-student architecture is adopted in~\cite{chen2022system} to learn visuomotor grasping policy.
These methods decouple visual knowledge and policy knowledge and learn in a two-stage fashion. However, they require accessing the underlying proprioceptive state as observation to train the teacher agent.

\noindent{\textbf{Pretraining.}}
Several works pretrain policies on offline demonstrations~\cite{zhan2020framework,wang2022vrl3}. However, the training data is from the policy learning environment and requires expert policy to collect trajectories, limiting the scalability and efficiency.
On the contrary, RRL~\cite{shah2021rrl} pretrains ResNet on ImageNet dataset as image encoder and freezes it in following policy learning process.
Xiao~\textit{et al.}~\cite{xiao2022masked} pretrain on in-the-wild frames via masked image modeling.
Our proposed ACO conducts pretraining based on YouTube videos which are captured in real world. We train an inverse dynamics model to create pseudo action labels and perform contrastive learning conditioned on these pseudo action labels.
As a concurrent work, VPT~\cite{baker2022video} similarly pretrains on online videos (game playing of Minecraft) and gets pseudo action labels from a trained inverse dynamics model. VPT pretrains by behavior cloning, different from contrastive learning used in our method.

\subsection{Contrastive Learning}

Contrastive learning~\cite{he2020momentum,wu2018unsupervised} is a popular pretext task for self-supervised learning. It creates supervisory labels via considering each image (instance) in the dataset forms a unique category and applies the learning objective of instance discrimination.
Previous works~\cite{he2020momentum,wu2018unsupervised,chen2020simple,grill2020bootstrap} in computer vision use contrastive learning to learn general vision knowledge that transfers well to various downstream tasks.
In reinforcement learning, CURL~\cite{laskin2020curl} and ATC~\cite{stooke2021decoupling} leverage contrastive learning as an additional signal to learn representation with environmental reward.

Common practice~\cite{he2020momentum} considers two views, $i.e.$ different augmentations, of a single image as a positive pair and views of different images as negative pairs. We call this type of positive/negative pair as Instance Contrastive Pair (ICP).
However, ICP only preserves knowledge for visual discriminability without task-relevant information.
We introduce Action Contrastive Pair (ACP) that contrasts images based on the underlying actions. Learning together with ACP and ICP will force the representation to focus on the visual elements most relevant to the decision-making process.
\section{Method}

\begin{figure}[t]
\centering
\includegraphics[width=\linewidth]{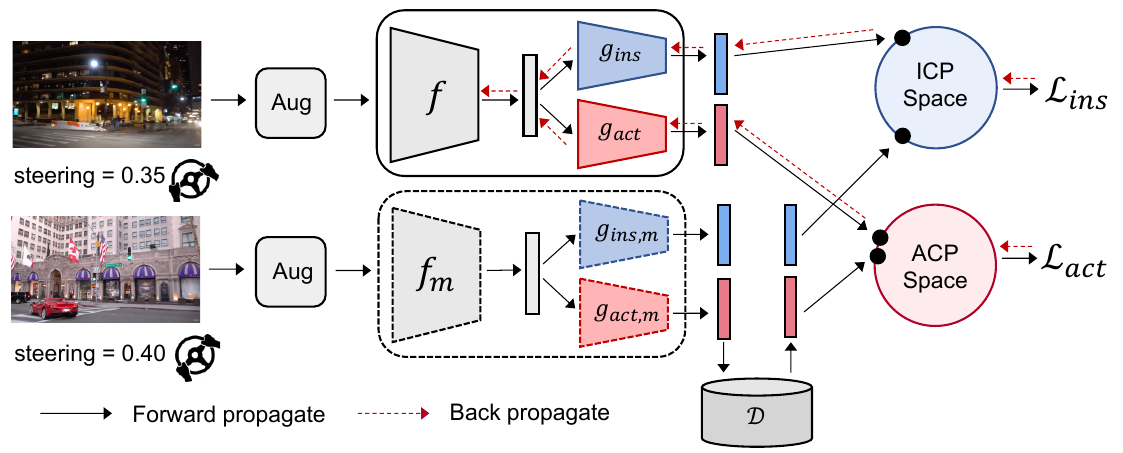}
\caption{
In the training pipeline of ACO, we form both ICP and ACP for the input images and update the model with contrastive losses in ICP space and ACP space.
The upper branch shows the calculation flow of query images and the lower shows the flow to generate the features of key images.
History key features $\mathcal D$ are stored in a key dictionary and re-used later
}
\label{figure:framework}
\end{figure}

We aim to learn generalizable visual feature for visuomotor policy learning, by pretraining on large amounts of uncurated real-world driving videos.
We propose a novel contrastive learning algorithm called \textbf{A}ction-Conditioned \textbf{CO}ntrastive Learning (\textbf{ACO}).

The essential of ACO is that we define two types of contrastive pair: Instance Contrastive Pair (ICP) and Action Contrastive Pair (ACP). Two views of a single image form a positive ICP, while two views of different images form a negative ICP. 
Only pretraining with ICP may make the representation include unnecessary information for downstream policy learning tasks. 
For example, visual cues like weather and lighting conditions are essential for forming ICP but contribute little to decision-making in autonomous driving.
To focus the feature on policy-relevant properties, we introduce another type of contrastive pair called Action Contrastive Pair (ACP).
Positive ACP is composed of two different images showing scenarios where the drivers' actions are similar. As an example, the two real-world snapshots in Figure~\ref{figure:framework} form a positive ACP, both showing the first-view images when the drivers are turning left.
ACO learns representation based on both ICP and ACP, where ICP focuses on learning discriminative general visual feature and ACP focuses on policy-relevant feature.

Figure~\ref{figure:framework} illustrates the training pipeline of ACO.
In upper branch, we first augment the given images twice to create query and key views and form ICPs.
As discussed in Section~\ref{section:ICP}, the ICPs are used to compute the ICP loss for visual discriminative feature learning.
We create another learning flow based on ACP. To create ACP, each frame in the dataset is tagged with a pseudo action label as introduced in Section~\ref{section:label}. In Section~\ref{section:ACP}, the action labels are used to form ACPs between different images and compute the ACP loss for policy feature learning.

\subsection{Visual Feature Learning with ICP}
\label{section:ICP}
As shown in Figure~\ref{figure:framework}, our contrastive learning is performed in ICP space and ACP space. To project given images into ICP space and form ICP, our pipeline contains the following parts:

\noindent
\textbf{Data Augmentation} module $Aug(\cdot)$. For each sample image $x$, we generate two random augmentations, $x^q=Aug(x)$ as query view and $x^k = Aug(x)$ as key view.

\noindent
\textbf{Encoder} module $f(\cdot)$. Each query view $x^q$ is further mapped to a feature vector $v^q=f(x^q)\in \mathcal{R}^{d_E}$, where $d_E$ is the feature dimension. For the key view $x^k$, a momentum-updated Encoder module $f_m(\cdot)$ is used to extract feature vector $v^k=f_m(x^k)\in \mathcal{R}^{d_E}$. The update rule of the encoder's parameters is:
\begin{equation}
    \theta_{f_m}\leftarrow \alpha \theta_{f_m} + (1-\alpha)\theta_{f},
    \label{eq:gmomentum1}
\end{equation}
where $\alpha$ is the momentum coefficient and $\theta_{f}$, $\theta_{f_m}$ are the parameters of $f$ and $f_m$, respectively.
The output of encoder module is the representation passed to policy head in downstream task.

\noindent
\textbf{Projector} module $g_{ins}(\cdot)$. 
We map key and query feature $v^q$ and $v^k$ to instance-contrastive space:
    \begin{equation}
        z_{ins}^q = g_{ins}(v^q),
        \quad\quad
        z_{ins}^k = g_{ins,m}(v^k), 
    \end{equation}
    where $z_{ins}^q, z_{ins}^k \in \mathcal{R}^{d_P}$ and $d_P$ is the dimension of the projected vector. $z_{ins}^q$ and $z_{ins}^k$ are further normalized to the unit hypersphere in $R^{d_P}$. 
    The projector $g_{ins,m}$  for key representation is also updated in a momentum way:
    \begin{equation}
        \theta_{g_{ins,m}} \leftarrow \alpha \theta_{g_{ins,m}} + (1-\alpha) \theta_{g_{ins}}.
        \label{eq:gmomentum2}
    \end{equation}
    
\noindent
\textbf{Key Dictionary} module $\mathcal D$. Following \cite{he2020momentum}, we use a key dictionary to store historical encoded key $z^k_{ins}$ to enable larger contrastive batchsize. 
Concretely, we generate key features $z^k$ of images following the lower branch in Figure~\ref{figure:framework}.
These key features not only form positive pairs with query features at current training epoch, but are also stored into $\mathcal D$ and used to form negative pairs at future training, if sampled.
The samples in the dictionary are progressively replaced if $\mathcal D$ exceeds its maximum size.

At each iteration of training, a batch of images are sampled from dataset.
The images are sequentially processed by Data Augmentation, Encoder and Projector modules to get query and key features.
The query-key pair forms positive ICP since they come from the same image.
To create negative ICP, we sample a batch (num=$N$) of key features from $\mathcal D$ and form negative pairs with current queries.
The negative keys are concatenated with current key features to form the key set $K$. 
Current key features are inserted into key dictionary for later use.
Then we perform contrastive learning on ICP between query and key features.

The ICP loss for visual feature learning is computed as:
\begin{equation}
\begin{aligned}
\mathcal{L}_{ins}&=
-\mathop{\log} \frac{
\sum_{z^+\in P_{ins}(z^q_{ins})}\exp(z_{ins}^q\cdot z^+ / \tau)
}{
\sum_{z^-\in N_{ins}(z^q_{ins})}\exp(z^q_{ins}\cdot z^- / \tau)
}, \\
P_{ins}(z^q)&=\{z~|~\text{id}(z)=\text{id}(z^q), z\in K \}, \\
N_{ins}(z^q)&\equiv K\backslash P_{ins}(z^q).
\end{aligned}
\end{equation}
The symbol $\cdot$ denotes inner (dot) product, $\text{id}(z^q) = x$ returns the image from which the projected vector is generated, $P_{ins}(z^q)$ and $N_{ins}(z^q)$ are the sets of keys of all positive and negative ICP, respectively.
Notice that $P_{ins}(z^q)$ only contains one sample and $N_{ins}(z^q)$ contains $N$ samples. This is because only one element in key set comes from the anchor image $x$ while other key features are sampled from $\mathcal D$.

\begin{figure}[t]
\centering
\includegraphics[width=1\linewidth]{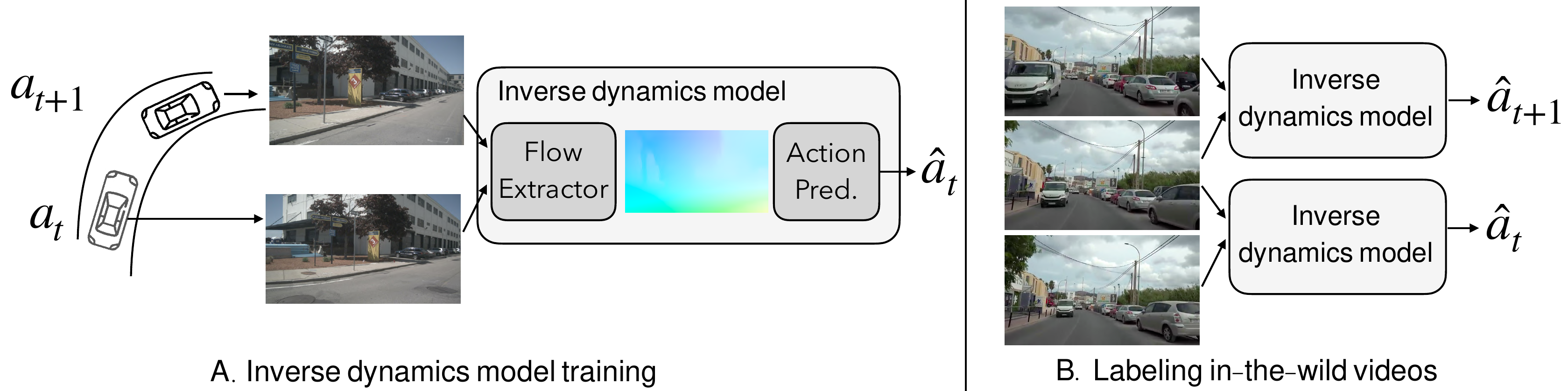}
\caption{
\textbf{A.} We use real-world driving dataset to train an inverse dynamics model. The inverse dynamics model is composed of a frozen flow extractor and action prediction head. 
\textbf{B.} We use the trained inverse dynamics model to label actions on large corpus of in-the-wild first view driving videos
}
\label{figure:inverse}
\end{figure}

\subsection{Generating Pseudo Labels for Video Frames}
\label{section:label}
We propose to conduct contrastive learning based on drivers' actions.
However, uncurated videos on the web do not contain any ground-truth action. 
We thus train an inverse dynamics model $\phi$ to predict the pseudo action label~\cite{kumar1905learning,torabi2018behavioral} $\hat{a}_t$ that has happened between two consecutive frames $I_{t}$ and $I_{t+1}$.
Nevertheless, many visual cues, such as weather and lighting, are irrelevant to action prediction.
Using two RGB frames as input may introduce unnecessary distractions to the inverse dynamics model.
Instead, we find that employing optical flow, the concise description of interframe movement, improves action prediction accuracy.

As illustrated in Figure~\ref{figure:inverse}\textbf{A}, we first extract the optical flow between frames by using an off-the-shelf algorithm RAFT~\cite{teed2020raft}. 
We then use the optical flow as the input to the inverse dynamics model $\phi$ to predict the action.
We choose NuScenes~\cite{caesar2020nuscenes} dataset which contains consecutive first-view driving scenes with labeled steering action to perform supervised learning of $\phi$. L1 loss between ground truth action $a_t$ from the dataset and predicted $\hat{a}_t$ is optimized. As shown in Figure~\ref{figure:inverse}\textbf{B}, we then use the extracted optical flow together with the learned $\phi$ to label action for each frame in YouTube videos. The pseudo action labels are further used to formulate ACPs to be introduced in the next section.

\subsection{Policy Feature Learning with ACP}
\label{section:ACP}
Based on the pseudo action labels, we conduct contrastive learning on ACP space to discover policy-relevant representation.
We share the Data Augmentation and Encoder modules with ICP learning in Section~\ref{section:ICP}.
During learning with ACP, the projector $g_{act}(\cdot)$ shares the same architecture as ICP's, but not the weights. A momentum-updated ACP projector $g_{act,m}(\cdot)$ is updated in a similar way as ICP projector in Equation~\ref{eq:gmomentum2}.
Thus the key and query vectors in action-contrastive space are computed as:
\begin{equation}
        z_{act}^q = g_{act}(v^q) = g_{act}(f(x^q)), 
        \quad\quad
        z_{act}^k = g_{act,m}(v^k) = g_{act,m}(f_m(x^k)).
\end{equation}

We also sample a batch of key features from $\mathcal D$ and concatenate them with current key features to form key set $K$.
The ACP loss is computed as:
\begin{equation}
\begin{aligned}
\mathcal{L}_{act}&=-\mathop{\log} \frac{
\sum_{z^+\in P_{act}(z^q_{act})}\exp(z_{act}^q\cdot z^+ / \tau)
}{
\sum_{z^-\in N_{act}(z^q_{act})}\exp(z^q_{act}\cdot z^- / \tau)
}.
\end{aligned}
\end{equation}
Different from ICP, the positive and negative keys in ACP are determined based on the action similarity.
The projected vectors of frames with similar underlying actions form positive pairs.
Therefore, when collecting the key set, we additionally store the action labels $\hat{a}$ with the key features $z^k$.
We use L1 distance to measure the similarity of two actions and the positive and negative key sets are determined as:
\begin{equation}
    \begin{aligned}
     P_{act}(z^q)&= \{ z~| \ \|\hat{a} - \hat{a}^q\| < \epsilon, (z, \hat{a})\in K \}, \\
     N_{act}(z^q)&\equiv K\backslash P_{act}(z^q),
    \end{aligned}
\end{equation}
where $\hat{a}^q$ is the predicted pseudo action label of the query image and $\epsilon$ is the action distance threshold. 

\subsection{Loss}

The overall contrastive loss of ACO is:
\begin{equation}
    \begin{aligned}
        \mathcal{L} = \lambda_{ins}\mathcal{L}_{ins} + \lambda_{act}\mathcal{L}_{act},
    \end{aligned}
    \label{eq:total-loss}
\end{equation}
where $\lambda_{ins}$ and $\lambda_{act}$ are weighted factors to balance ICP and ACP.

\section{Experiments}

We evaluate the pretrained feature from our method on two main policy learning tasks: Imitation Learning (IL) and Reinforcement Learning (RL) in end-to-end driving. We also evaluate on one related visual recognition task: Lane Detection (LD).
Section~\ref{section:baselines} lists the compared baselines.
The main experimental results with a comparison of different pretraining methods are presented in Section~\ref{section:il} for IL, in Section~\ref{section:rl} for RL, and in Section~\ref{section:lanedet} for LD. 
Section~\ref{section:visualization} gives a qualitative analysis of the learned action-conditioned representation.
Ablation study and further discussion are included in Section~\ref{section:ablation}.

\noindent
\textbf{YouTube driving dataset.}
We crawl first-view driving videos from YouTube. 134 videos with a total length of over 120 hours are collected. As shown in Figure~\ref{figure:teasor}, these videos cover different driving scenes with various weather conditions (sunny, rainy, snowy, etc.) and regions (rural and urban areas). We sample one frame every one second, resulting to a dataset of 1.30 million frames.
We split the YouTube driving dataset into training set with 70\% data and test set with 30\% data and conduct the training of ACO on the training set.

\begin{figure}[t]
\centering
\includegraphics[width=\linewidth]{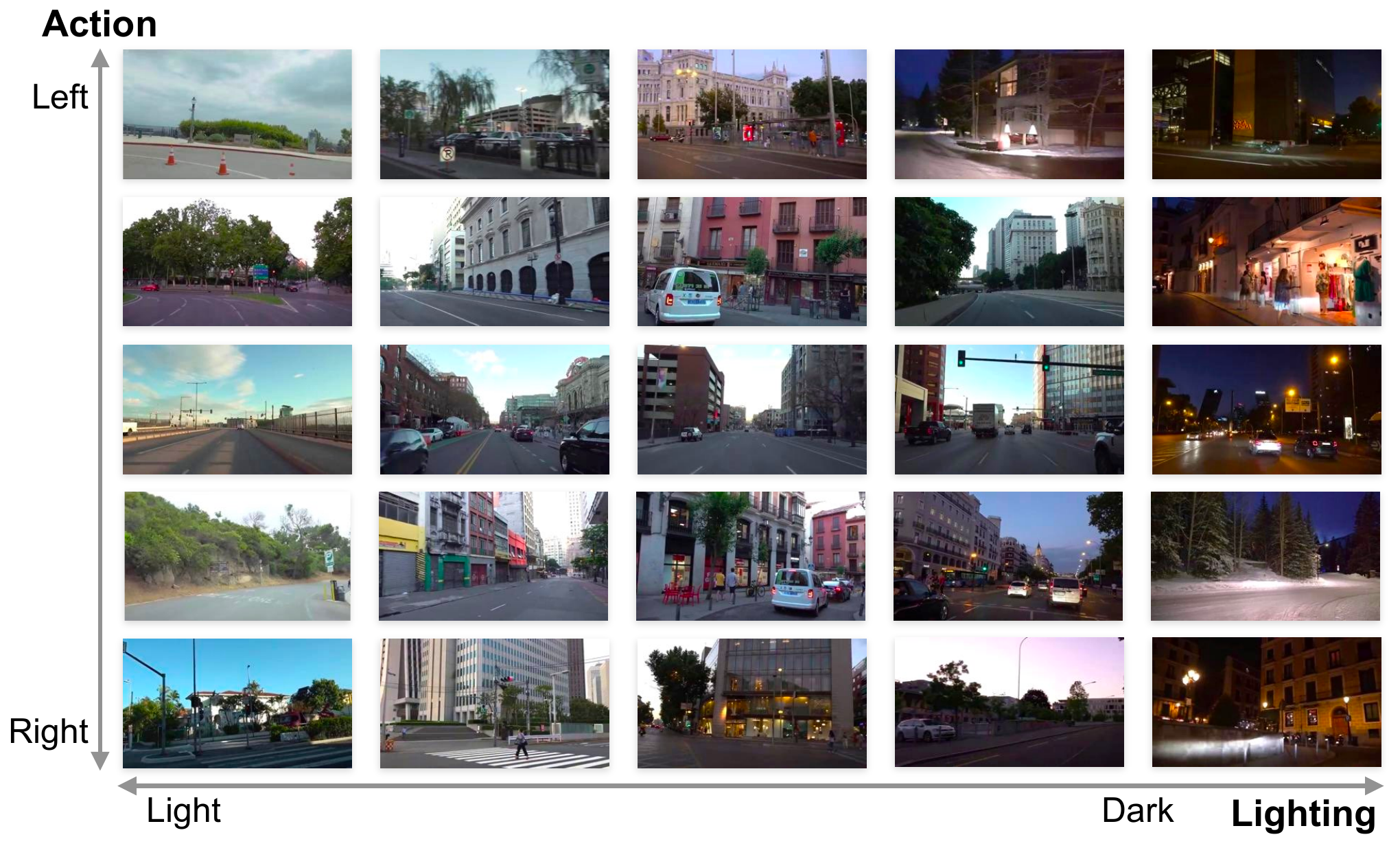}
\caption{
\textbf{Exemplar video frames organized by their pseudo labels.} We use the inverse dynamics model to label each frame's action. The predicted action is mostly correct despite diverse visual variations of the image content
}
\label{fig:inverselabel}
\end{figure}

\noindent
\textbf{Inverse dynamics model.}
An inverse dynamics model with ResNet-34 architecture is used to predict action labels for frames in YouTube driving dataset.
We use RAFT~\cite{teed2020raft} as the optical flow extractor and apply Adam optimizer with learning rate of 0.001 and weight decay of 0.001 to learn the model on the training set of NuScenes~\cite{caesar2020nuscenes}.
We normalize steering value between 0 and 1.
In our preliminary experiments, the L1 error of steering prediction averaged over the test set of NuScenes~\cite{caesar2020nuscenes} is 0.10 when using consecutive images as input, and 0.04 when using optical flow.
Using optical flow as the input brings better inverse dynamics prediction performance as it is more invariant to the image content. 
Figure~\ref{fig:inverselabel} shows some sampled frames. The steering action is predicted accurately under different visual conditions.

\noindent
\textbf{Pretraining.}
In the representation learning, we largely follow the hyper-parameters from the official implementation of MoCo-v2~\cite{chen2020improved}. We optimize the model with synchronized SGD over 8 GPUs with a weight decay of 0.0001, a momentum of 0.9, and a batch size of 32 on each GPU. We perform 100 epochs optimization with an initial learning rate of 0.03 and a cosine learning rate schedule. Two projectors for ICP and ACP are both instantiated as two-layer MLPs. We maintain a memory queue of 40,960 samples as the Key Dictionary $\mathcal D$. The momentum coefficient in Equation~\ref{eq:gmomentum1} and Equation~\ref{eq:gmomentum2} is set to 0.999 for updating the key encoder and projector.

\noindent
\textbf{Data augmentations.}
During pretraining, we follow MoCo-v2~\cite{chen2020improved} and use image augmentations to generate key and query views of an image. Specifically, we apply random gray scaling, color jittering and Gaussian blurring. Random resized cropping and horizontal flipping are excluded because we find that they will largely change the scene's semantic, thus degrading the overall performance, as shown in the ablation study. 

\subsection{Baselines}
\label{section:baselines}

We choose ResNet-34 as the backbone (encoder) architecture and compare the downstream tasks' performance when using different pretrained models as the initial parameters of the backbone network. Different pretraining methods are as follows:

\noindent
\textbf{Random.} The backbone's parameters are initialized randomly. 
Specifically, we use Kaiming normalization~\cite{he2015delving} to initialize convolution layers.
Constant initialization is applied to batchnorm layer (1 for weight and 0 for bias).

\noindent
\textbf{ImageNet.} We use official ImageNet's pretrain weights to initialize the backbone. This is considered as the most common approach for current image-based policy learning~\cite{shah2021rrl}.

\noindent
\textbf{AutoEncoder.} We train an autoencoder on the collected YouTube videos and use the encoder to initialize the backbone, which has been used in previous policy learning work~\cite{ha2018world}.

\noindent
\textbf{MoCo.} As a straightforward comparison to our method, contrastive learning is conducted following MoCo-v2~\cite{chen2020improved} on the collected YouTube videos. 
As a baseline, we only use ICP to conduct contrastive learning and set $\lambda_{act} = 0$ in Equation~\ref{eq:total-loss}.

\begin{table}[!t]
\caption{\textbf{Success Rate of Imitation Learning.} We evaluate the performance of different pretrained models under IL tasks with different scales of dataset
}
\label{tab:il}
\centering
\setlength{\tabcolsep}{12pt}
\begin{tabular}{@{}ccccc@{}}
\toprule
\multicolumn{1}{c}{\multirow{2}{*}{Pretrain Method}} & \multicolumn{4}{c}{IL Demonstration Size ($\times 40K$)} \\ \cmidrule(l){2-5} 
\multicolumn{1}{c}{}                                 & 10\%     & 20\%    & 40\%    & 100\%      \\ \midrule
Random                                               & 0.0$\pm$0.0       & 0.0$\pm$0.0      & 37.3$\pm$6.6     & 81.3$\pm$5.2     \\
AutoEncoder                                          & 0.0$\pm$0.0       & 4.0$\pm$3.3      & 6.7$\pm$2.5     & 46.0$\pm$5.9     \\
ImageNet                                             & 21.3$\pm$7.5      & 52.7$\pm$13.1     & 72.0$\pm$4.3     & 90.7$\pm$7.4   \\
MoCo                                                 & 19.3$\pm$3.4                &    60.3$\pm$8.2       &    76.0$\pm$9.1    &   80.7$\pm$2.5      \\ \midrule
\textbf{ACO}                                                 & \textbf{30.7}$\pm$3.4      & \textbf{66.0}$\pm$5.7     & 
\textbf{82.0}$\pm$5.0    
& \textbf{96.0}$\pm$3.3    \\ \bottomrule
\end{tabular}
\end{table}

\subsection{Imitation Learning}
\label{section:il}

\noindent
\textbf{Setup.}
We evaluate the performance of different pretrained weights for Imitation Learning in the open-source CARLA~\cite{DBLP:journals/corr/abs-1711-03938} simulator.
We use the original CARLA~\cite{DBLP:journals/corr/abs-1711-03938} benchmark (also known as CORL2017) and collect 50 trajectories of expert demonstration, which amount to 40K transitions, in Town01 with train weather. Town02 with test weather is used for evaluation.
The expert is a rule-based PID controller with injected noise~\cite{codevilla2018end}. 
We adopt CILRS~\cite{codevilla2019exploring}, a conditional behavior clonning algorithm, to train policies based on different pretrained models.
The backbone network is fine-tuned during policy learning. We also conduct IL with frozen backbone and conclude the result in Appendix. All experiments are repeated three times with different random seeds. Information about other hyper-parameters is given in Appendix.

\noindent
\textbf{Results.}
To see how pretrained models improve IL tasks, we experiment on the expert datasets with 10\%, 20\%, 40\% and 100\% of all 40K samples.
As shown in Table~\ref{tab:il}, ACO outperforms all other baselines in datasets of all scales.
Notably, AutoEncoder is less competitive even compared to random initialization. This is because pixel reconstruction target is misaligned with policy learning target and makes the encoder pay more attention to areas with larger size like buildings and the sky, which however are useless for imitation learning. MoCo and ImageNet perform well at full dataset but have degraded performance when dataset size is reduced. ACO-pretrained models consistently outperform baselines at all different dataset sizes.

\begin{figure}[!t]
\centering
\includegraphics[width=\linewidth]{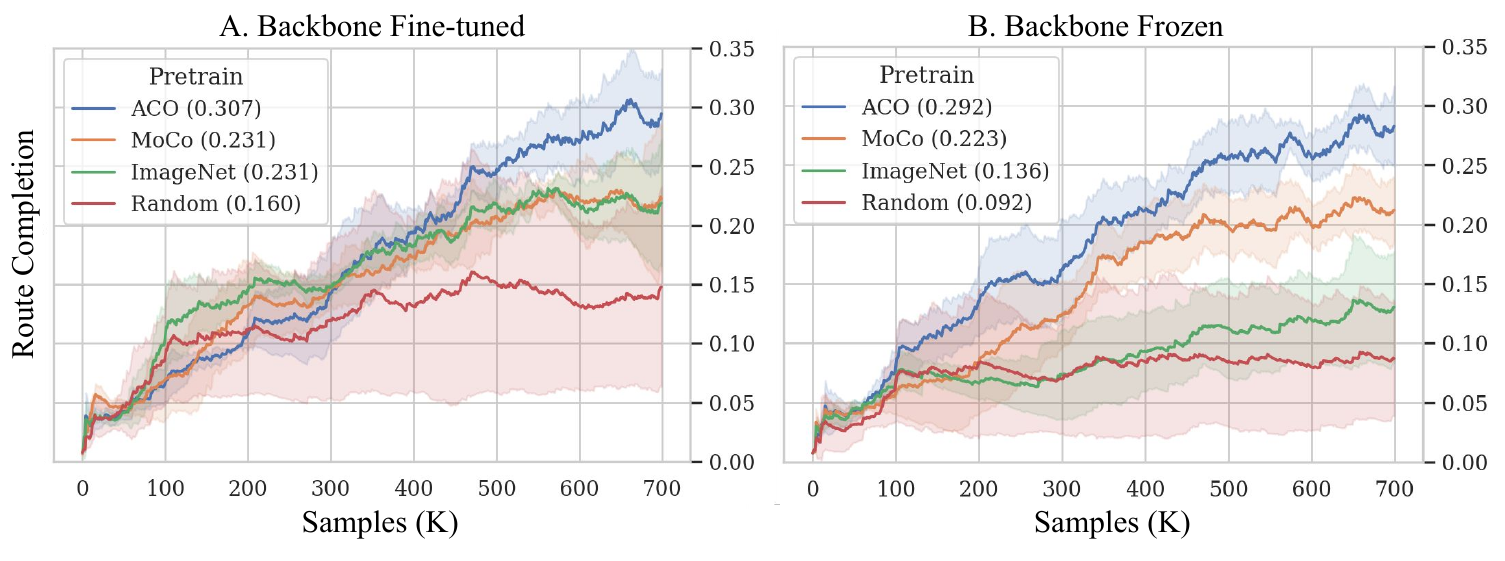}
\caption{
\textbf{Route completion curves for PPO~\cite{schulman2017proximal} with different pretrain weights.} In the left panel, the backbone is actively updated during policy learning. In the right panel, the backbone is frozen and not updated. The proposed ACO outperforms all baselines
}
\label{fig:rl}
\end{figure}

\subsection{Reinforcement Learning}
\label{section:rl}

\noindent
\textbf{Setup.}
Similar to Imitation Learning, we also use CARLA~\cite{DBLP:journals/corr/abs-1711-03938} as the simulator for evaluating pretrained models in Reinforcement Learning.
We use Town01 in NoCrash benchmark~\cite{codevilla2019exploring} as the environment and evaluate the performance of Proximal Policy Optimization (PPO)~\cite{schulman2017proximal} with different pretrained weights.

\noindent
\textbf{Results.}
As shown in Figure~\ref{fig:rl}\textbf{A},
PPO with ACO initialization introduces a clear improvement of 7\% route completion compared to other baselines when the backbone is fine-tuned during policy learning. This suggests that incorporating policy-relevant objects into pretraining stage benefits downstream RL's performance. ImageNet and MoCo, which only learn visual knowledge, are less competitive compared to ACO. For random initialization, we observe a poor 15\% route completion, in which case the agent can only drive in straight road but fails to learn turning.

To further understand the importance of considering policy-relevant information in pretraining, we freeze the backbone and only finetune the policy head during RL.
As shown in Figure~\ref{fig:rl}\textbf{B}, compared to backbone fine-tuned case, the performance of PPO declines dramatically when using ImageNet pretraining. 
This indicates that ImageNet pretrained weight lacks the essential knowledge for policy learning thus leads to a poor RL performance. ACO's and MoCo's performance are not affected when freezing the backbone. 

\subsection{Lane Detection}
\label{section:lanedet}

\noindent
\textbf{Setup.}
To further prove that the representation learned by ACO is generalizable, we conduct lane detection experiments.
We train three popular lane detection methods on CULane dataset~\cite{pan2018SCNN}: UFLD~\cite{qin2020ultra}, SCNN~\cite{pan2018SCNN}, and RESA~\cite{zheng2020resa}. All of these methods employ backbone networks with ResNet architecture. We use ACO and other baselines to initialize the backbone's parameters before training.
We adopt the training settings (\textit{e.g.} learning rate) of three LD methods fine-tuned on ImageNet pretrained backbone. Better performance is expected after tunning the hyper-parameters for different pretrained methods respectively.
MoCo~\cite{he2020momentum} suggests that a batch normalization layer after backbone can alleviate the gap between contrastive learning and supervised learning caused by different parameters' distribution. 
We thus follow this procedure on ACO and MoCo baseline and present the results with ``-bn'' suffix.

\noindent
\textbf{Results.}
As shown in Table~\ref{tab:lanedet}, ImageNet pretraining achieves the best performance among all candidates. Although trained in an object-centric dataset, ImageNet still provides high quality feature for lane detection learning. 
Within the self-supervised baselines, ACO outperforms others in UFLD~\cite{qin2020ultra} and RESA~\cite{zheng2020resa}.
In SCNN~\cite{pan2018SCNN}, ACO is on par with MoCo.
Introducing batch normalization improves the performance of ACO.

\begin{table}[!t]
\caption{\textbf{F1 metric of lane detection}}
\setlength{\tabcolsep}{8pt}
\label{tab:lanedet}
\centering
\begin{tabular}{@{}cccccc@{}}
\toprule
\multicolumn{1}{c}{\multirow{2}{*}{Pretrain Method}} & \multicolumn{3}{c}{LD Method}             \\ \cmidrule(l){2-4} 
\multicolumn{1}{c}{}                          & UFLD~\cite{qin2020ultra} & SCNN~\cite{pan2018SCNN} & RESA~\cite{zheng2020resa} \\ \midrule 
\textit{ImageNet}                                      & \textit{70.9} & \textit{74.1} & \textit{76.1}        \\ \midrule
Random                                        & 68.4 & 70.4 & 73.9     \\
AutoEncoder                                            & 68.6 & 70.1 & 74.0     \\
MoCo                                          & 69.5 & \textbf{72.0} & 74.1        \\
MoCo-bn                                          & 69.8 & 71.9 & 74.1        \\ \midrule
ACO                                          & 69.5 & 71.6 & 74.5   \\ 
ACO-bn                                          & \textbf{70.4} & 71.8 & \textbf{75.0}   \\ \bottomrule
\end{tabular}
\end{table}

\subsection{Visualization}
\label{section:visualization}

We perform t-SNE~\cite{van2008visualizing} analysis on features extracted by different layers of ACO and compare them with that of MoCo.
We randomly select frames from the test set and divide them into three categories: \textit{left turn}, \textit{right turn}, and \textit{go straight}.
Each category has 25 examples. Images within same category form positive ACPs with one another.
To build positive ICPs, we sample the previous and next frame of chosen samples and consider them as positive pairs.

\begin{figure}[!t]
\centering
\includegraphics[width=\linewidth]{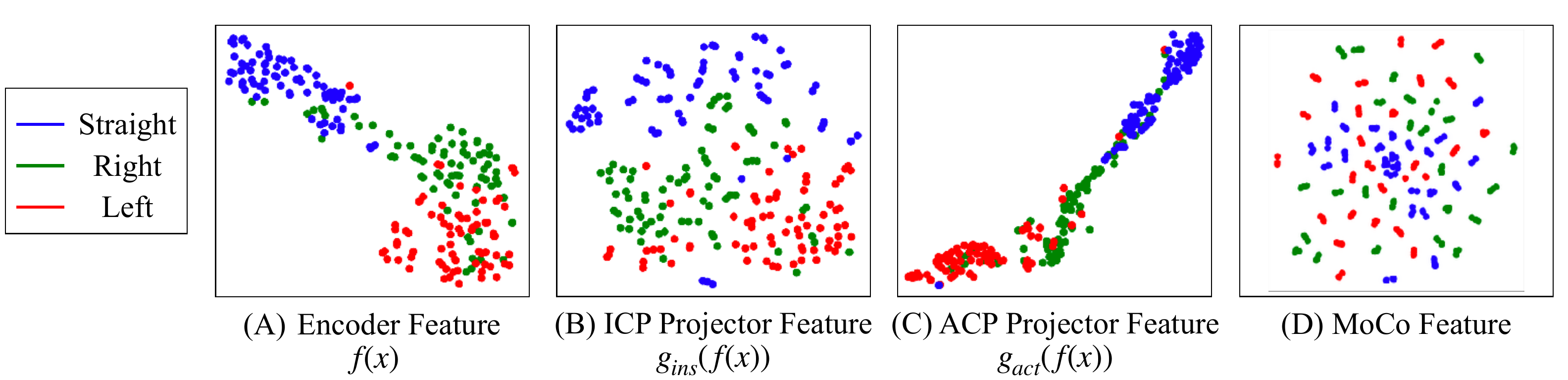}
\caption{
\textbf{t-SNE~\cite{van2008visualizing} visualization of extracted features.}
Red, green, and blue dots indicate frames under \textit{left turn}, \textit{right turn}, and \textit{go straight} scenarios respectively
} 
\label{fig:tsne}
\end{figure}

As shown in Figure~\ref{fig:tsne}\textbf{A}, features extracted by ACO's encoder, which will be passed to policy head in downstream task, form separable clusters and have clear semantics related to the possible actions of drivers.
On the contrary, the action-conditioned clustering does not emerge in MoCo in Figure~\ref{fig:tsne}\textbf{D}. Instead, we find that MoCo feature demonstrates the \textit{instance clustering phenomenon}, where the features of three temporal neighboring frames lay closely and form a clique in the t-SNE visualization.

Diving into the contrastive learning modules, we find the ICP projector demonstrates instance clustering as in Figure~\ref{fig:tsne}\textbf{B}.
The phenomenon does not exist in ACP projected features in Figure~\ref{fig:tsne}\textbf{C}, where features cluster more tightly according to action information instead of temporal relationship.
t-SNE visualization shows that ICP and ACP in ACO both achieve their respective goals: ICP endows features with the capacity to discriminate across instances, whereas ACP creates features that tightly couple with action information.

\begin{table}[!t]
\caption{\textbf{Ablation study of pretraining datasets.}
We use ACO to pretrain models on our YouTube dataset and the NuScenes dataset respectively, and conduct IL training on CARLA
}
\setlength{\tabcolsep}{5pt}
\label{tab:ablation-dataset}
\centering
\begin{tabular}{@{}cccccc@{}}
\toprule
\multirow{2}{*}{
\shortstack[b]{Pretraining\\Dataset}
}
&
\multirow{2}{*}{Frame num (M)} 
&
\multicolumn{4}{c}{IL Demonstration Size ($\times 40K$)} \\ \cmidrule(l){3-6} 
\multicolumn{1}{c}{}                    &             & 10\%     & 20\%    & 40\%    & 100\%      \\ \midrule
NuScenes~\cite{caesar2020nuscenes}                                &   0.19$\MVAt$10Hz &   13.3$\pm$3.4       & 27.3$\pm$10.4      & 67.3$\pm$9.0     & 82.7$\pm$8.4     \\
\textbf{YouTube}                           &      \textbf{1.30}$\MVAt$1Hz              & \textbf{30.7}$\pm$3.4      & \textbf{66.0}$\pm$5.7     & 
\textbf{82.0}$\pm$5.0    
& \textbf{96.0}$\pm$3.3    \\ \bottomrule
\end{tabular}
\end{table}

\begin{table}[!t]
\caption{\textbf{Ablation study of augmentations.} CF stands for Cropping and Flipping}
\setlength{\tabcolsep}{5pt}
\label{tab:ablation-augmentation}
\centering
\begin{small}
\begin{tabular}{@{}lcccc@{}}
\toprule
\multicolumn{1}{l}{\multirow{2}{*}{Methods}} & \multicolumn{4}{c}{IL Demonstration Size ($\times 40K$)} \\ \cmidrule(l){2-5} 
\multicolumn{1}{c}{}                        & 10\%     & 20\%    & 40\%    & 100\%      \\ \midrule
MoCo w/ CF                   &  19.3$\pm$3.4                &    60.3$\pm$8.2       &    76.0$\pm$9.1    &   80.7$\pm$2.5           \\
MoCo w/o CF                                  &    13.3$\pm$3.4      & 47.3$\pm$13.7     & 72.7$\pm$9.0     & 84.0$\pm$4.9   \\ \midrule
ACO w/ CF                                  &   27.3$\pm$0.9      &  36.0$\pm$5.9    &  74.0$\pm$7.5    &  89.3$\pm$5.0     \\
ACO w/o CF                                   & \textbf{30.7}$\pm$3.4      & \textbf{66.0}$\pm$5.7     & 
\textbf{82.0}$\pm$5.0    
& \textbf{96.0}$\pm$3.3    \\ \bottomrule
\end{tabular}
\end{small}
\end{table}

\subsection{Ablation Study}
\label{section:ablation}

We conduct ablation studies that examine the effectiveness of collecting and labeling YouTube driving videos rather than using NuScenes dataset~\cite{caesar2020nuscenes} and the impact of geometric-aware augmentations (cropping and flipping).

\noindent
\textbf{YouTube Dataset.}
To highlight the importance of collecting large dataset from YouTube, we train IL agents based on models pretrained on different datasets and compare the success rate on IL benchmark.
As shown in Table~\ref{tab:ablation-dataset}, the size of our collected YouTube dataset is six times larger than NuScenes dataset~\cite{caesar2020nuscenes} even though we sample frames at 1Hz from the video, compared to 10Hz of NuScenes.
The imitation learning performance of ACO trained on YouTube dataset outperforms that on NuScenes dataset~\cite{caesar2020nuscenes} at all IL training settings. 
The result shows that YouTube dataset improves the generalizability of the pretrained models and leads to better performance in the downstream task.

\noindent
\textbf{Augmentations.}
We conduct imitation learning experiments to discover the impact of two augmentations, cropping and flipping frames (CF in the following for lucidity).
As shown in Table~\ref{tab:ablation-augmentation}, the performance of MoCo without CF is significantly lower than MoCo with these two augmentations, showing that CF is important for instance discriminative learning. 
However, the result suggests ACO is incompatible with CF. Excluding CF in ACO will affect the feature's instance discriminative ability but improve the overall performance in downstream driving tasks with action-conditioned contrastive learning.
\section{Conclusion and Discussion}
In this paper, we propose a novel contrastive policy pretraining method ACO using hours-long uncurated YouTube videos.
By learning action-conditioned features from unlabeled video frames with pseudo action labels, our methods  greatly improve the generalizability of the learned representation and brings substantial improvements to  downstream tasks.

\vspace{0.5em}
\noindent
\textbf{Limitations.}
Despite the generalizable representation provided by pretraining with action information, we incorporate a strong assumption: only one action corresponds to each driving scene in a single video frame. This does not hold true as drivers may have various driving intentions in complex scenarios. Future study will concentrate on how to label frames more accurately, \textit{e.g.} labeling each frame with an action distribution, to produce more precise action contrastive pairs.
Besides, action-conditioned contrastive learning has conflict with widely used geometric-aware augmentations like cropping and flipping, suppressing feature's instance discrimination ability. We leave this problem for future study.

\clearpage
%
%
\bibliographystyle{splncs04}
\bibliography{egbib}
\clearpage

\appendix
\setcounter{footnote}{0}

\section{YouTube Driving Dataset}

\begin{figure}[]
\centering
\includegraphics[width=0.7\linewidth]{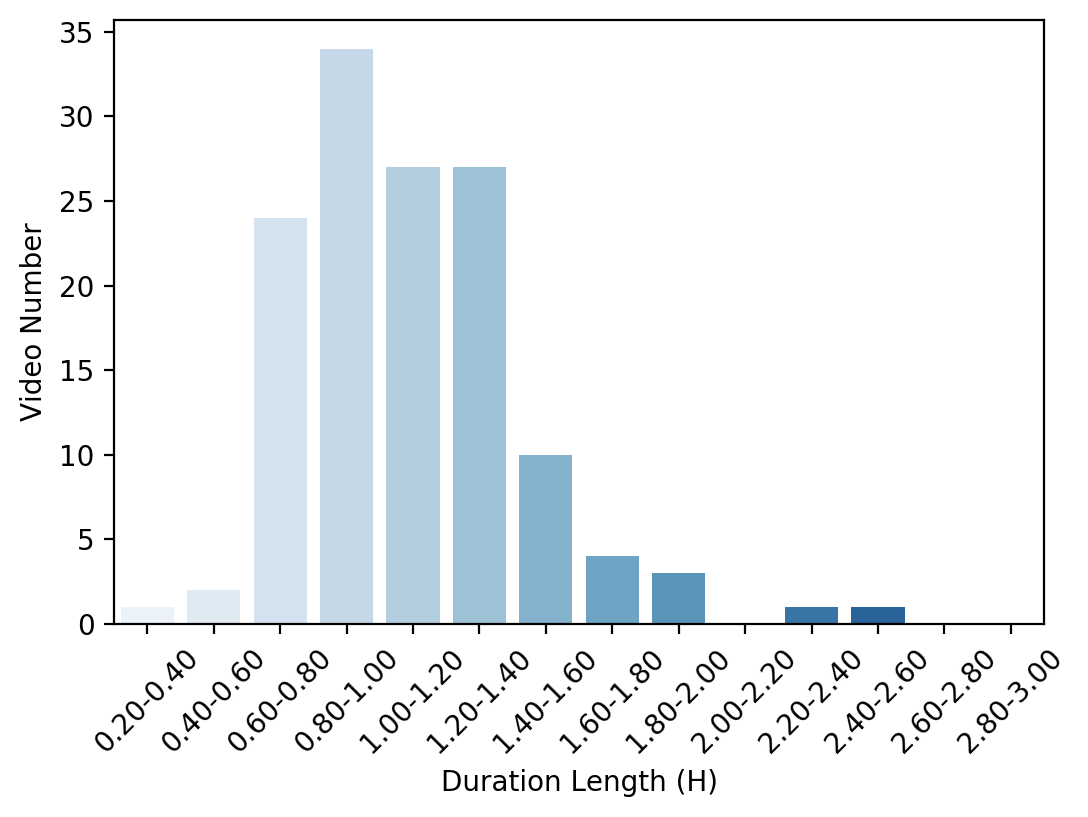}
\caption{\textbf{Video length distribution}}
\label{fig:supp-stat}
\end{figure}

Our YouTube Driving Dataset contains a massive amount of real-world driving frames with various conditions, from different weather, different regions, to diverse scene types.
We collect 134 videos with a total length of over 120 hours, covering up to 68 cities. Figure~\ref{fig:supp-stat} illustrates the videos' lengths distribution.
For pretraining, we sample one frame every one second, resulting to a dataset of 1.30 million frames in total.

\begin{table}[h]
\caption{\textbf{Success Rate of Imitation Learning with frozen backbone}}
\setlength{\tabcolsep}{5pt}
\label{tab:fil}
\centering
\begin{tabular}{@{}lcccc@{}}
\toprule
\multicolumn{1}{c}{\multirow{2}{*}{Methods}} & \multicolumn{4}{c}{IL Demonstration Size ($\times 40K$)} \\ \cmidrule(l){2-5} 
\multicolumn{1}{c}{}                                 & 10\%     & 20\%    & 40\%    & 100\%      \\ \midrule
Random-f                                               & 0.0$\pm$0.0       & 0.0$\pm$0.0      & 2.0$\pm$0.0     & 4.0$\pm$0.0     \\
ImageNet-f                                             & 4.0$\pm$0.0      & 4.0$\pm$2.8     & 5.3$\pm$0.9     & 6.7$\pm$0.9   \\
MoCo-f             & 10.7$\pm$2.5 &  30.7$\pm$4.1   & 38.7$\pm$1.9 & 54.7$\pm$6.6  \\ \midrule
\textbf{ACO-f}     & \textbf{24.0}$\pm$1.6      & \textbf{40.7}$\pm$3.4     & 
\textbf{47.3}$\pm$0.9    
& \textbf{61.3}$\pm$3.8    \\ \bottomrule
\end{tabular}

\end{table}

\begin{table}[]
\caption{\textbf{Effects of Batch Normalization in Imitation Learning.} ``bn'' stands for Batch Normalization}
\setlength{\tabcolsep}{5pt}
\label{tab:supp-bn}
\centering
\begin{small}
\begin{tabular}{@{}lcccc@{}}
\toprule
\multicolumn{1}{l}{\multirow{2}{*}{Methods}} & \multicolumn{4}{c}{IL Demonstration Size ($\times 40K$)} \\ \cmidrule(l){2-5} 
\multicolumn{1}{c}{}                        & 10\%     & 20\%    & 40\%    & 100\%      \\ \midrule
MoCo w/o bn                   &  17.3$\pm$1.9                &    55.3$\pm$16.8       &    59.3$\pm$5.7    &   80.0$\pm$1.6           \\
MoCo w/ bn                                  &    \textbf{19.3}$\pm$3.4                &    \textbf{60.3}$\pm$8.2       &    \textbf{76.0}$\pm$9.1    &   \textbf{80.7}$\pm$2.5    \\ \midrule
ACO w/o bn                                  &   27.3$\pm$3.8      &  34.0$\pm$4.3    &  72.0$\pm$4.3    &  78.0$\pm$4.3     \\
ACO w/ bn                                   & \textbf{30.7}$\pm$3.4      & \textbf{66.0}$\pm$5.7     & 
\textbf{82.0}$\pm$5.0    
& \textbf{96.0}$\pm$3.3    \\ \bottomrule
\end{tabular}
\end{small}
\end{table}

\section{Imitation Learning with frozen backbone}

As shown in Table~\ref{tab:fil}, we conduct imitation learning with frozen backbone.
With backbone frozen, ACO demonstrates highest success rate. 
However, all methods suffer from severe performance drop compared to finetuned IL.
For ImageNet and random pretrain, they drop by over 80\% in full dataset size. 

In contrast, RL methods exhibit less performance drop with frozen backbone.
As shown in Fig.~5 in main paper, the biggest drop by frozen RL compared to finetuned RL is less than 50\%.
This phenomenon shows that freezing the backbone exacerbates the need for a good pretrain for IL.
Policy trained with bad pretrain weight will give suboptimal actions, for example, driving the car to the road edge. IL, notoriously suffering from \textit{distribution shift}, has no means to recover this error since the collected demonstrations do not contain any recovery action. Nevertheless, RL frequently visits dangerous states during exploration, and recovery strategies can hence be learned. 

\section{Effects of Batch Normalization on ACO}

As suggested in \cite{he2020momentum}, Batch Normalization could help alleviate the problem caused by different distributions of contrastive-learning-based parameters and supervised-learning-based parameters.
In this section, we compare the imitation learning performance on ACO and MoCo with and without batch normalization in downstream IL's fine-tuning.

As shown in Table~\ref{tab:supp-bn}, Batch Normalization is essential for both MoCo and ACO. Without adjusting parameters' distribution by batch normalization, these two contrastive learning methods show declining performance.


\begin{figure}[]
\centering
\includegraphics[width=\linewidth]{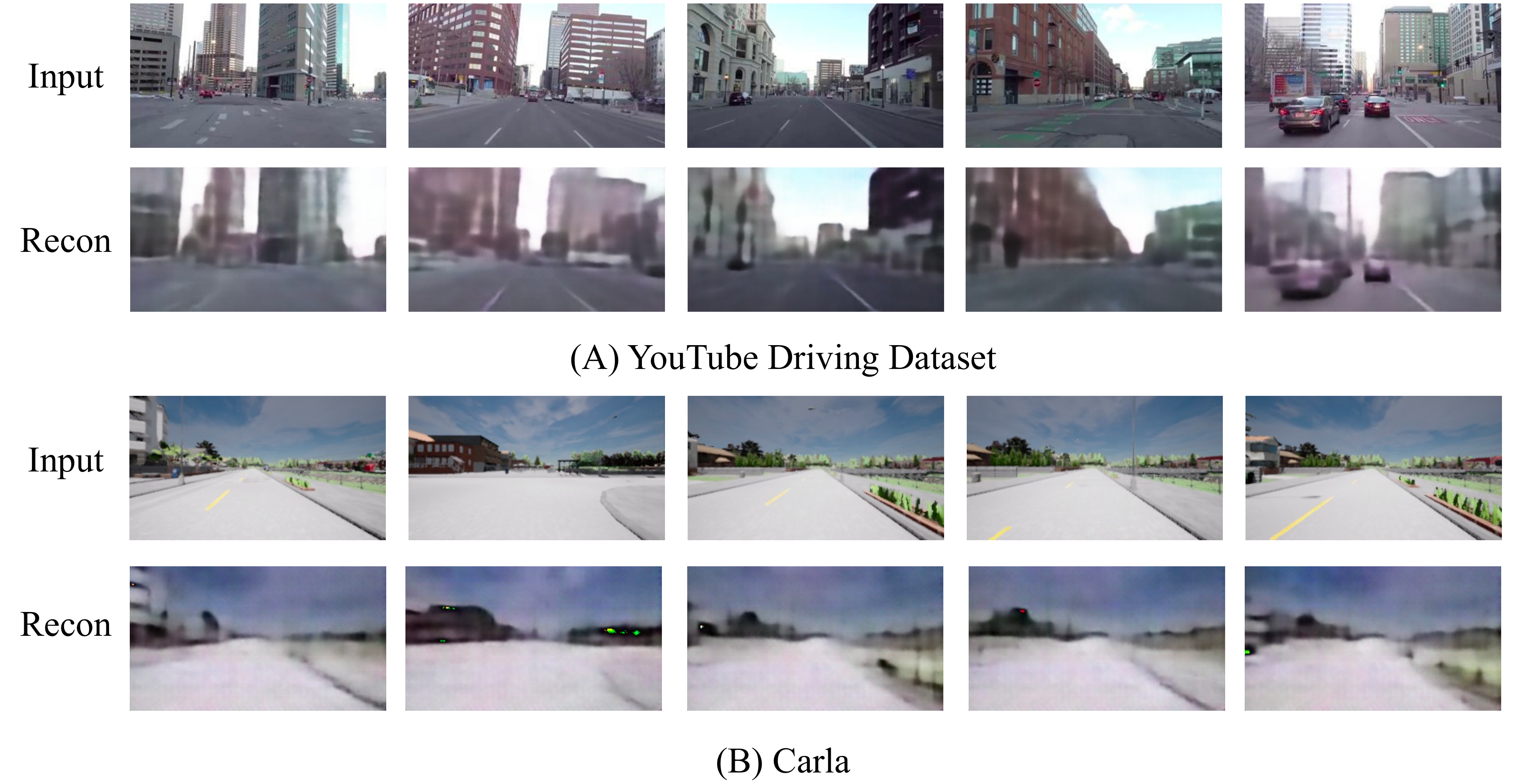}
\caption{\textbf{Reconstructed scenes by AutoEncoder in YouTube and Carla domain}}
\label{fig:supp-ae}
\end{figure}

\section{Visualization of Auto-Encoder Baseline}

To understand the poor performance of auto-encoder in all downstream tasks, we visualize the input and reconstructed image in both YouTube (training) and Carla (testing) domain in Figure~\ref{fig:supp-ae}.

It is obvious that much attention is paid to elements that are not relevant to driving decision-making in reconstructed images, such as buildings and the sky.
On the contrary, the policy-relevant cues such as the lane lines and road edges are oversimplified. 
This means that useless information is learned in auto-encoder's feature and thus damages the performance of downstream tasks.

\section{Implementation Details and Hyper Parameters}

We list the hyper-parameters used in ACO pretraining in Table~\ref{tab:supp-hyper-aco}, inverse dynamics model training in Table~\ref{tab:supp-hyper-inv}, imitation learning in Table~\ref{tab:supp-hyper-il}, and reinforcement learning in Table~\ref{tab:supp-hyper-rl}.

For imitation learning, following CILRS~\cite{codevilla2019exploring}, the training target is to optimize L1loss between predicted value and ground truth on steering, throttle, brake, and velocity:
\begin{equation}
    \mathcal{L}_{il} =  \lambda_{s}\mathcal{L}_{s}+ \lambda_{t}\mathcal{L}_{t} +\lambda_{b}\mathcal{L}_{b}+\lambda_{v} \mathcal{L}_{v},
\end{equation}
where $\mathcal{L}_{s}$, $\mathcal{L}_{t}$, $\mathcal{L}_{b}$, and $\mathcal{L}_{v}$ stands for L1loss on steering, throttle, brake, and velocity respectively.
The value of $\lambda_{s}$,  $\lambda_{t}$,  $\lambda_{b}$, and  $\lambda_{v}$ we used  are listed in Table~\ref{tab:supp-hyper-il}.

For reinforcement learning, we adopt the training pipeline in DI-drive~\cite{didrive}\footnote{DI-drive~\cite{didrive} is an open-source auto-driving platform, supporting reinforcement learning in Carla. The reward scheme can be found at: \url{https://github.com/opendilab/DI-drive/blob/main/core/envs/simple\_carla\_env.py}.}.

\begin{table}[]
\centering
\caption{ACO}
\begin{tabular}{@{}ll@{}}
\toprule
Hyper-parameter                          & Value  \\ \midrule
Initial learning rate                    & 0.03   \\
Weight decay                             & 0.0001 \\
Train epoch number                       & 100    \\
Train batch size                         & 256    \\
Key dictionary size                      & 40960  \\
$\lambda_{ins}$                                   & 1      \\
$\lambda_{act}$                          & 1      \\
$\alpha$ for momentum update.            & 0.999  \\
Hidden size of first layer in Projector  & 128    \\
Hidden size of second layer in Projector & 128    \\
$\epsilon$ for action threshold              & 0.05   \\ \bottomrule
\end{tabular}
\label{tab:supp-hyper-aco}
\end{table}

\begin{table}[]
\centering
\caption{Inverse dynamics model}
\setlength{\tabcolsep}{12pt}
\begin{tabular}{@{}ll@{}}
\toprule
Hyper-parameter                          & Value  \\ \midrule
Learning rate                    & 0.0003   \\
Weight decay                             & 0.0001 \\
Train epoch number                       & 50    \\
Train batch size                         & 256    \\\bottomrule
\end{tabular}
\label{tab:supp-hyper-inv}
\end{table}

\begin{table}[]
\centering
\setlength{\tabcolsep}{12pt}
\caption{Imitation learning (CILRS~\cite{codevilla2019exploring})}
\begin{tabular}{@{}ll@{}}
\toprule
Hyper-parameter                          & Value  \\ \midrule
Learning rate                    & 0.0001   \\
Weight decay                             & 0.0001 \\
Train epoch number                       & 100    \\
Train batch size                         & 128    \\
Velocity loss weight  $\lambda_{v}$                        & 0.05    \\
Steering loss weight  $\lambda_{s}$                      & 0.5    \\
Throttle loss weight    $\lambda_{t}$                     & 0.45    \\
Brake loss weight     $\lambda_{b}$                      & 0.05  \\\bottomrule
\end{tabular}
\label{tab:supp-hyper-il}
\end{table}

\begin{table}[]
\centering
\setlength{\tabcolsep}{12pt}
\caption{Reinforcement learning (PPO~\cite{schulman2017proximal})}
\begin{tabular}{@{}ll@{}}
\toprule
Hyper-parameter                          & Value  \\ \midrule
Learning rate                    & 0.0003   \\
Weight decay                             & 0.00001 \\
SGD epoch number                       & 6    \\
$\lambda$ for GAE~\cite{schulman2015high}                       & 0.95    \\
Train batch size                         & 1024    \\
SGD mini batch size                         & 256    \\
Clip ratio                                  & 0.2   \\ \bottomrule
\end{tabular}
\label{tab:supp-hyper-rl}
\end{table}

\section{Sampled frames of YouTube Driving Datasets}

\begin{figure}[]
\centering
\includegraphics[width=\linewidth]{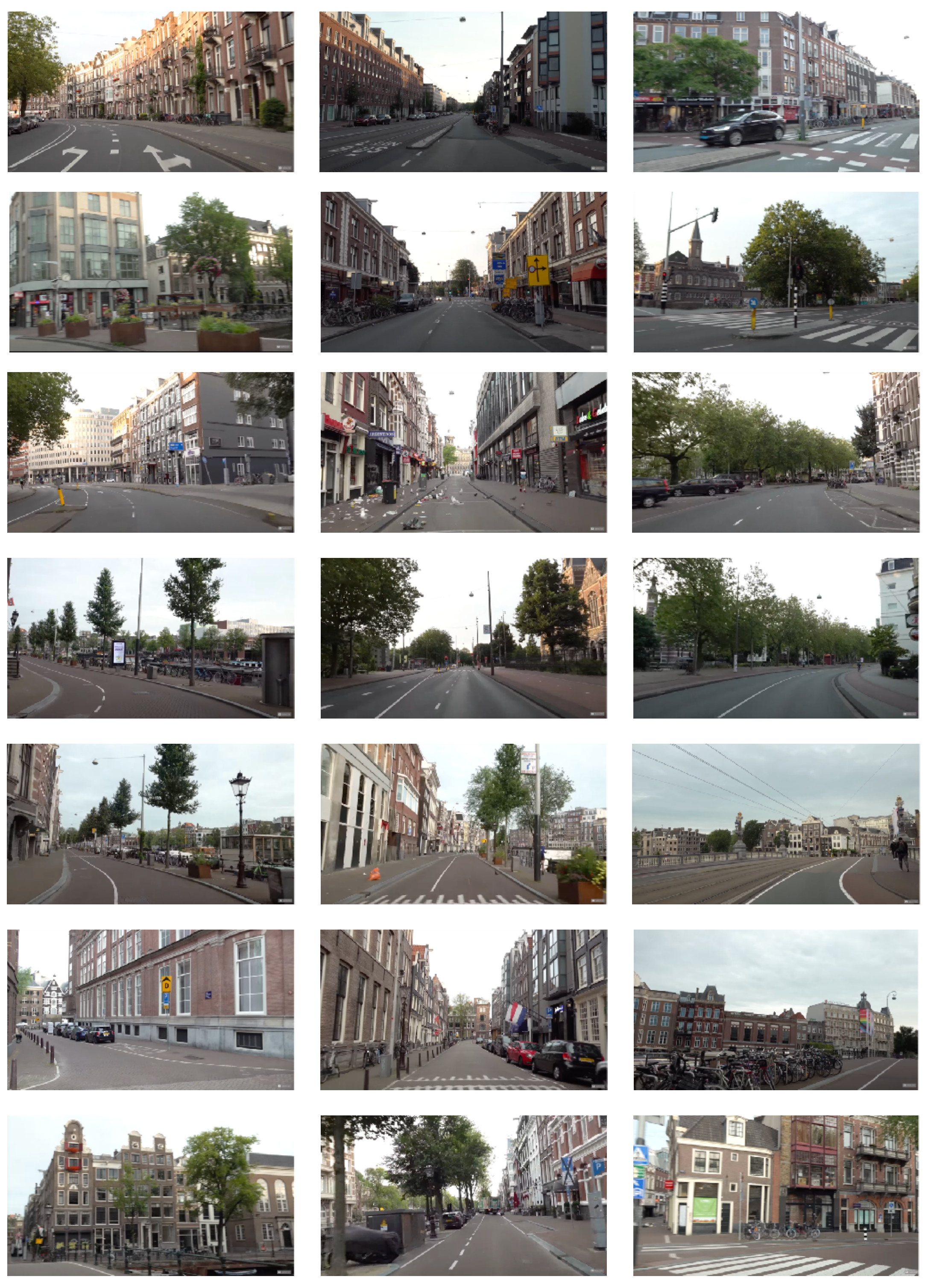}
\caption{\weather{Early morning} drive at the \region{center of Amsterdam},
    \scenetype{empty traffic} on \scenetype{straight roads}, with \scenetype{urban buildings} surrounded}
\label{fig:supp-frame-1}
\end{figure}

\begin{figure}[]
\centering
\includegraphics[width=\linewidth]{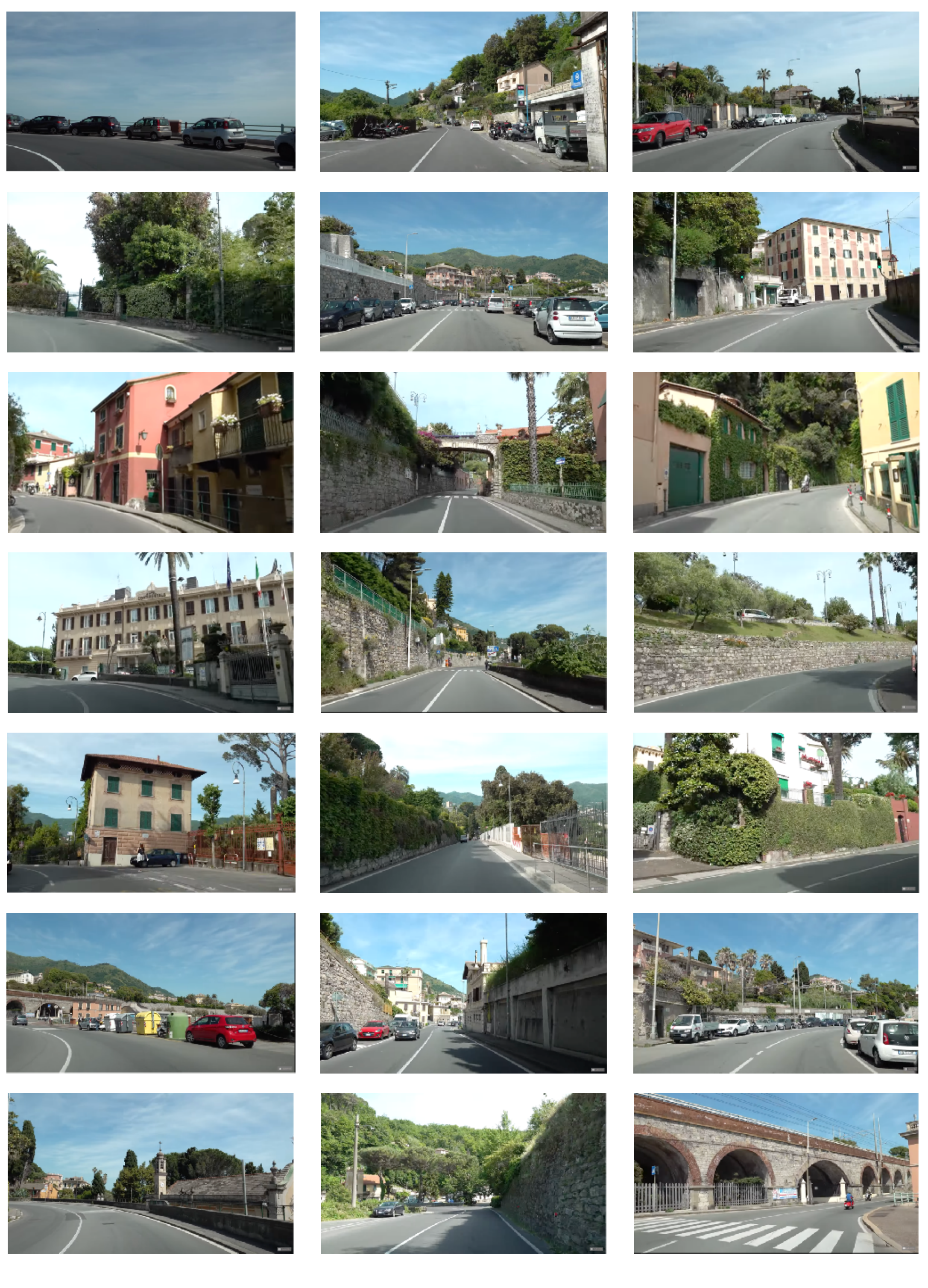}
\caption{\weather{Afternoon} drive along the \region{Italian Riviera},           
     along \scenetype{narrow roads} with \scenetype{frequent turnings}}
\label{fig:supp-frame-2}
\end{figure}

\begin{figure}[]
\centering
\includegraphics[width=\linewidth]{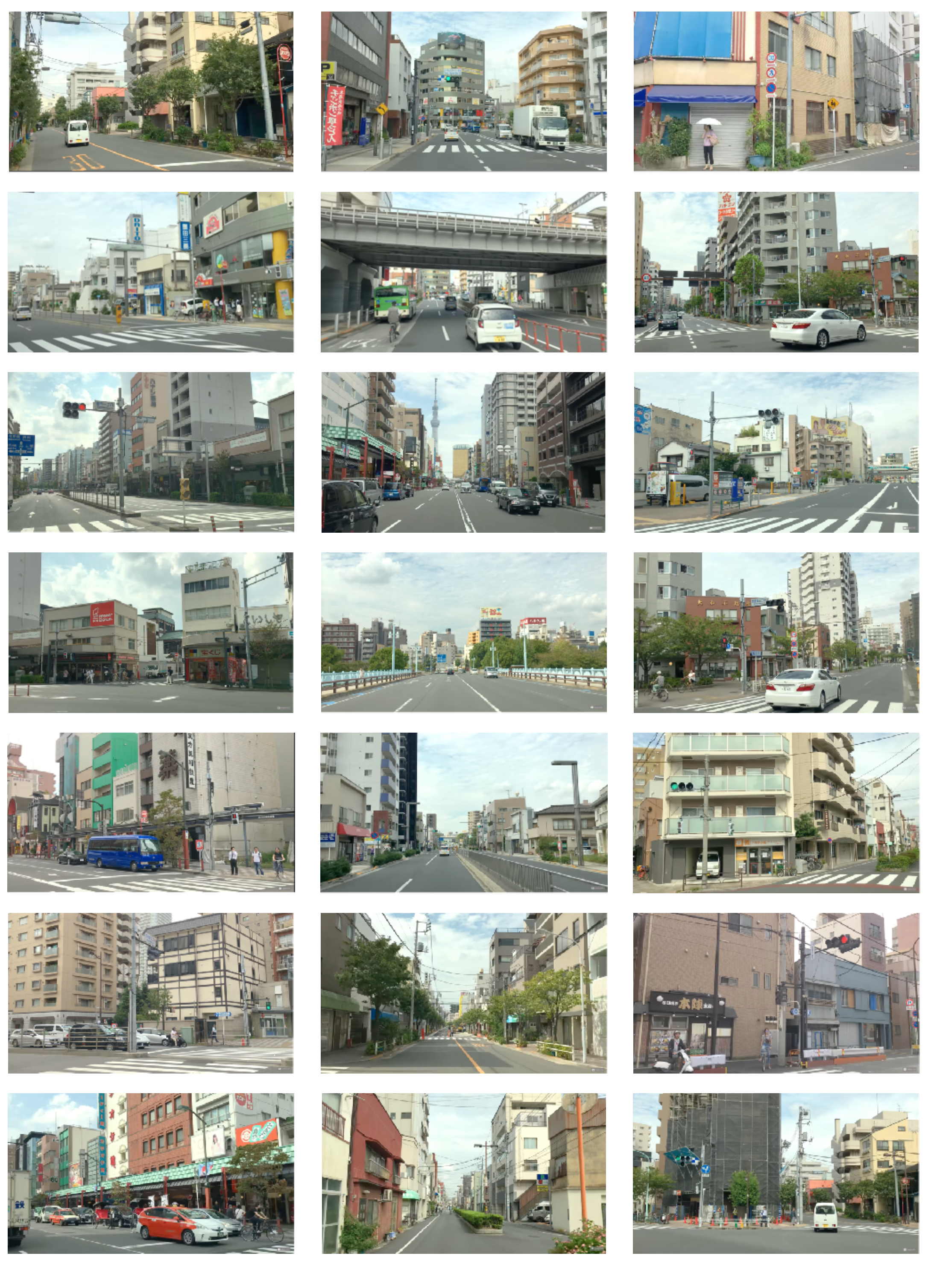}
\caption{\weather{Afternoon} drive around the traditional \region{Asakusa} neighborhood, 
      through \scenetype{rich traditional streets}. \scenetype{Left-hand} riding with \scenetype{middle density traffic}}
\label{fig:supp-frame-3}
\end{figure}

\begin{figure}[]
\centering
\includegraphics[width=\linewidth]{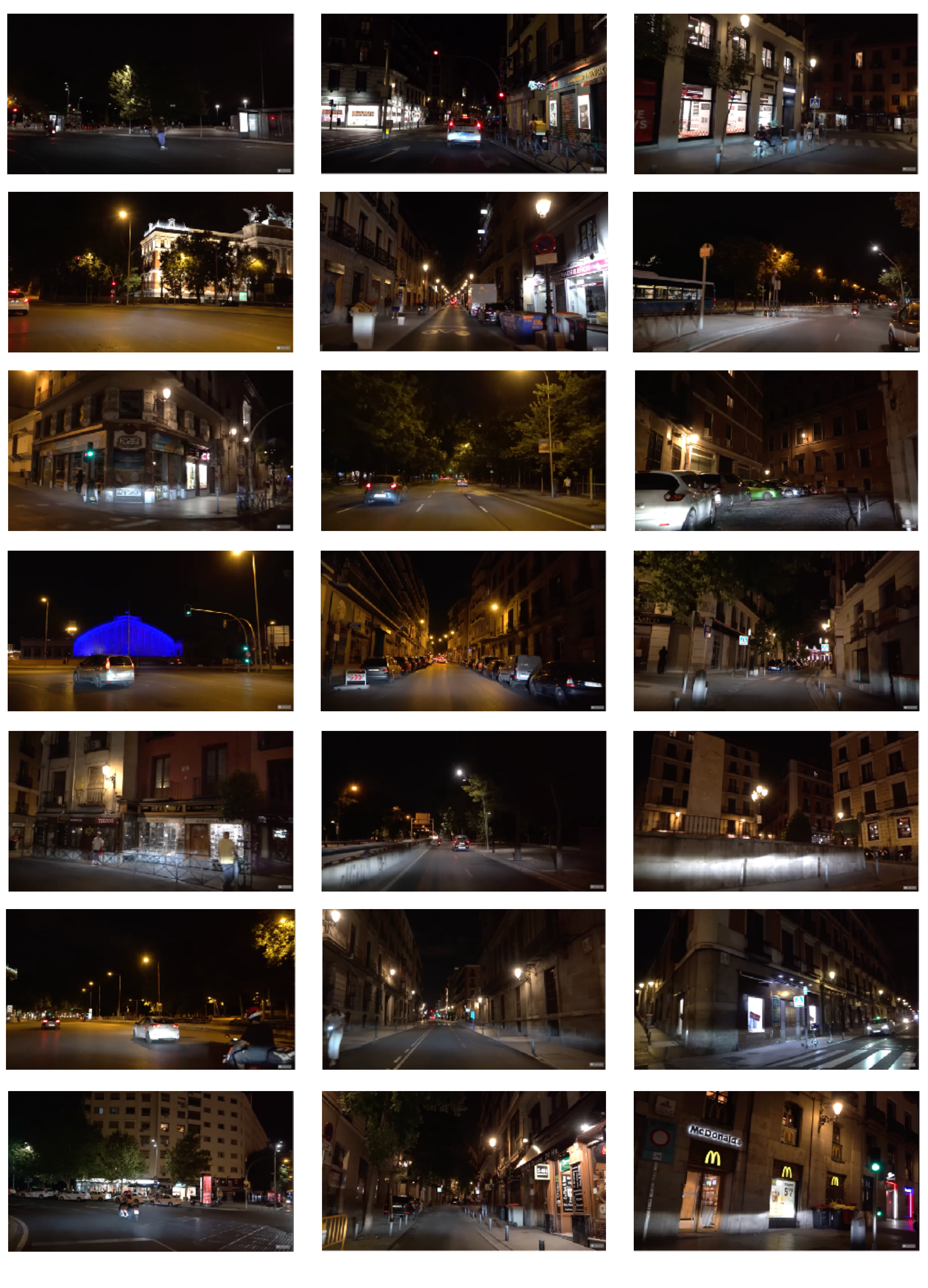}
\caption{\weather{Night drive} in \region{Madrid Spain},         
       through \scenetype{Straight roads} with \scenetype{middle density traffic}}
\label{fig:supp-frame-4}
\end{figure}

\begin{figure}[]
\centering
\includegraphics[width=\linewidth]{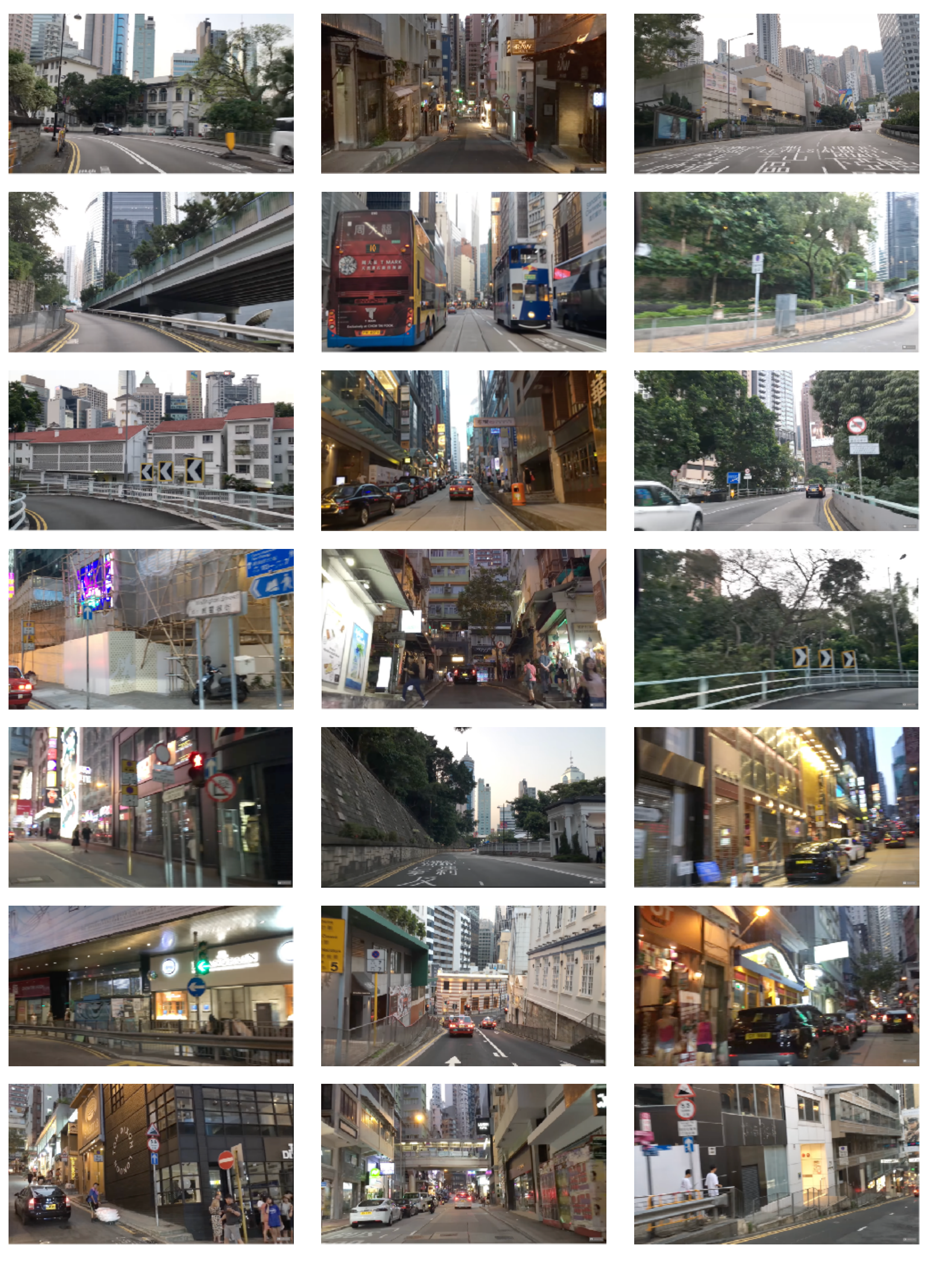}
\caption{\weather{Sunset drive} in \region{Lan Kwai Fong neighborhood, Hong Kong}, 
       \scenetype{middle traffic, modern building, left-hand, some one-way streets}}
\label{fig:supp-frame-5}
\end{figure}

We provide sampled frames of five cities in Figure~\ref{fig:supp-frame-1},~\ref{fig:supp-frame-2},~\ref{fig:supp-frame-3},~\ref{fig:supp-frame-4},~\ref{fig:supp-frame-5}.
Note that in captions we describe corresponding weather with \weather{blue text}, region with \region{red text}, and scene style with \scenetype{green text}. The columns are organized by different predicted steering  values.

\end{document}